\documentclass[a4paper]{article}

\usepackage[a4paper,top=3cm,bottom=2cm,left=3cm,right=3cm,marginparwidth=1.75cm]{geometry}

\usepackage{amsmath}
\usepackage{graphicx}
\usepackage[colorinlistoftodos]{todonotes}
\usepackage[colorlinks=true, allcolors=blue]{hyperref}
\usepackage{mathtools}
\usepackage{amssymb}
\usepackage{amsthm} 
\usepackage{tikz,graphics,color,float,epsf,caption,subcaption}

\newcommand{\enum}[1]{\label{eq:#1}}

\newcommand{\eref}[1]{(\ref {eq:#1})}

\newcommand{\be}{\begin{equation}}
\newcommand{\ee}{\end{equation}}

\newcommand{\bear}{\begin{eqnarray}}
\newcommand{\eear}{\end{eqnarray}}

\newcommand{\erefs}[2]{({\ref{eq:#1}}~--~{\ref{eq:#2}})}

\newcommand{\nc}{\newcommand}
\nc{\I}{\mathbf{I}}
\nc{\F}{\mathbf{F}}
\nc{\cc}{\mathbf{c}}
\nc{\dd}{\mathbf{d}}
\nc{\bC}{\mathbf{C}}
\nc{\bp}{\mathbf{p}}
\nc{\bL}{\mathbf{L}}
\nc{\Q}{\mathbf{Q}}
\nc{\B}{\mathbf{B}}
\nc{\W}{\mathbf{W}}
\nc{\Z}{\mathbf{Z}}
\nc{\N}{\mathbf{N}}
\nc{\M}{\mathbf{M}}
\nc{\h}{\mathbf{H}}
\nc{\bH}{\mathbf{H}}
\nc{\n}{\mathbf{n}}
\nc{\mpi}{\bm{\pi}}

\nc{\bP}{\mathbf{P}}

\nc{\D}{\mathbf{D}}
\nc{\G}{\mathbf{G}}
\nc{\bu}{\mathbf{u}}
\nc{\bbv}{\mathbf{v}}
\nc{\bw}{\mathbf{w}}
\nc{\bbf}{\mathbf{f}}

\nc{\A}{\mathbf{A}}
\nc{\bE}{\mathbf{E}}
\nc{\g}{\mathbf{g}}

\nc{\bde}{\boldsymbol{\delta}}
\nc{\bD}{\boldsymbol{\Delta}}
\nc{\ex}{\cal E}

\newtheorem{prop}{Proposition}

\usepackage[utf8]{inputenc} 
\usepackage[T1]{fontenc}    
\usepackage{hyperref}       
\usepackage{url}            
\usepackage{booktabs}       
\usepackage{amsfonts}       
\usepackage{nicefrac}       
\usepackage{microtype}      

\title{On weight initialization in deep neural networks}
\author{Siddharth Krishna Kumar}
\author{
  Siddharth Krishna Kumar \\
  \texttt{siddharthkumar@upwork.com}
}

\begin{document}

\maketitle

\begin{abstract}
A proper initialization of the weights in a neural network is critical to its convergence. Current insights into weight initialization come primarily from linear activation functions. In this paper, I develop a theory for weight initializations with non-linear activations.
First, I derive a general weight initialization strategy for any neural network using activation functions differentiable at 0. Next, I derive the weight initialization strategy for the Rectified Linear Unit (RELU), and provide theoretical insights into why the {\it{Xavier initialization}} is a poor choice with RELU activations. My analysis provides a clear demonstration of the role of non-linearities in determining the proper weight initializations.
\end{abstract}

\section{Introduction}

In recent years, there have been rapid advances in our understanding of deep neural networks. These advances have resulted in breakthroughs in several fields, ranging from image recognition (\cite{russakovsky2015imagenet},\cite{szegedy2013deep},\cite{toshev2014deeppose}) to speech recognition (\cite{graves2013speech},\cite{maas2013rectifier},\cite{sutskever2014sequence}) to natural language processing (\cite{collobert2008unified},\cite{kim2014convolutional}, \cite{socher2011parsing}). These successes have been achieved despite the notorious difficulty in training these deep models.

Part of the difficulty in training these models lies in determining the proper initialization strategy for the parameters in the model. It is well known \cite{mishkin2015all} that arbitrary initializations can slow down or even completely stall the convergence process. The slowdown arises because arbitrary initializations can result in the deeper layers receiving inputs with small variances, which in turn slows down back propagation, and retards the overall convergence process. Weight initialization is an area of active research, and numerous methods (\cite{mishkin2015all}, \cite{saxe2013exact}, \cite{sussillo2014random} to state a few) have been proposed to deal with the problem of the shrinking variance in the deeper layers.

In this paper, I revisit the oldest, and most widely used approach to the problem with the goal of resolving some of the unanswered theoretical questions which remain in the literature. The problem can be stated as follows: If the weights in a neural network are initialized using samples from a normal distribution, ${\cal{N}}(0,v^{2})$, how should  $v^{2}$ be chosen to ensure that the variance of the outputs from the different layers are approximately the same?


The first systematic analysis of this problem was conducted by Glorot and Bengio \cite{glorot2010understanding} who showed that for a linear activation function, the optimal value of $v^{2} = 1/N$, where $N$ is the number of nodes feeding into that layer. Although the paper makes several assumptions about the inputs to the model, it works extremely well in many cases and is widely used in the initialization of neural networks to date; this initialization scheme is commonly referred to as the  {\it{Xavier initialization}}.

In an important follow up paper, He and colleagues \cite{he2015delving} argue that the {\it{Xavier initialization}} does not work well with the RELU activation function, and instead propose an initialization of $v^{2} = 2/N$ (commonly referred to as the {\it{He initialization}}). In support of their initialization, they provide an example of a 30 layer neural network which converges with the {\it{He initialization}}, but not under the {\it{Xavier initialization}}. To the best of my knowledge, the precise reason for the convergence of one method and the non-convergence of the other is not fully understood.

My main contributions in this paper are to (a) generalize the results of \cite{glorot2010understanding} to the case of non-linear activation functions and (b) to provide a continuum between the results of \cite{glorot2010understanding} and \cite{he2015delving}. For the class of activation functions differentiable at 0, I provide a general initialization strategy. For the class of activation functions that are not differentiable at 0, I focus on the Rectified Linear Unit (RELU) and provide a rigorous proof of the {\it{He initialization}}. I also provide theoretical insights into why the 30 layer neural network converges with the {\it{He initialization}} but not with the {\it{Xavier initialization}}. As a small bonus, I resolve an unanswered question posed in \cite{glorot2010understanding} regarding the distributions of activations under the hyperbolic tangent activation. 

\section{The setup}
Consider a deep neural network with $M$ layers. The relationship between the inputs to the $m^{th}$ layer ($x_{m}$) and $m+1^{th}$ layer ($x_{m+1}$) are described by the recursions 
\be 
y_{m}(i) = \sum\limits_{j=1}^{j=N}{\bf{W}}_{m}(i,j)x_{m}(j) = \sum\limits_{j=1}^{j=N} p_{ij} \enum{in1}
\ee 
and 
\be 
x_{m+1}(i) = \mbox{g}(y_{m}(i)). \enum{in2} 
\ee 
Here  $p_{j} = {\bf{W}}_{m}(i,j)x_{m}(j)$, ${\bf{W}}_{m}$ is a matrix of weights for the $m^{th}$ layer,  $g$ is the non-linear activation function, and $N$ is the number of nodes in the hidden layers respectively. The weights ${\bf{W}}_{m}(i,j)$ are assumed to be independent identically distributed normal random variables with mean 0 and variance $v^{2}$. Consistent with the assumptions in \cite{glorot2010understanding} and \cite{he2015delving}, I assume that the inputs to the first layer are independent and identically distributed random variables with mean 0 and variance 1. For convenience, I use 
$r_{m}$ and $u_{m}^{2}$ to denote the mean and variance of $y_{m}(i)$ respectively.

Due to the symmetry in the problem, all inputs to the $m^{th}$ layer will have the same means and variances during the first forward pass (i.e., $E(x_{m}(i)) = \mu_m$ and  $Var(x_{m}(i)) = s_{m}^{2}$ for all $i$); the covariances between the inputs to the $m^{th}$ layer need not be 0.

The goal is to find the value of $v^{2}$  which ensures that $s_{1}^{2} \approx s_{2}^{2} ...\approx s_{M}^{2} = 1$ during the first forward pass. To accomplish this, I need to express the  central moments of $x_{m+1}(i)$ in terms of the central moments of $x_{m}(i)$ for an arbitrary value of $m$. I begin by analyzing properties of the neural network that are independent of the activation function considered in the analysis.
\begin{prop}
During the first iteration, $W_{m}(i,j)$ is independent of $x_{m}(k)$ for all values of $i, j$ and $k$ 
\end{prop}
Using the recursions in \eref{in1} and \eref{in2}, $x_{m}(k)$ can be expressed as some non-linear function of the weights in the first $m-1$ layers, and the inputs to the first layer. Since the weights in the $m^{th}$ layer are independent of the inputs to the first layer and the weights in all other layers, the weights in the $m^{th}$ layer will also be independent of any non-linear function of these quantities. Therefore, $W_{m}(i,j)$ is independent of $x_{m}(k)$ for all values of $i, j$ and $k$. Furthermore, since $W_{m}(i,j)$ is independent of $x_{m}(k)$ and $x_{m}(l)$, $W_{m}(i,j)$ will also be independent of $x_{m}(k)x_{m}(l)$\qedsymbol
\linebreak
\newline
Taking expectations in \eref{in1} and using proposition 1, along with the fact that $E({\bf{W}}_{m}(i,j)) = 0$ yields 
\be 
r_{m} = E(y_{m}(i)) = \sum\limits_{j=1}^{j=N} E({\bf{W}}_{m}(i,j))E(x_{m}(j)) = 0. \enum{in201}
\ee 
Therefore,  $u_{m}^{2} = Var(y_{m}(i)) = E\left(y_{m}(i)^{2}\right) -  \left(E(y_{m}(i))\right)^{2} = E\left(y_{m}(i)^{2}\right)$. Using \eref{in1},
\be 
\begin{split}
u_{m}^{2} &= E\left(\sum\limits_{j=1}^{j=N}{\bf{W}}_{m}(i,j)x_{m}(j)\right)^{2} \\
&= \sum\limits_{j=1}^{j=N} E\left({\bf{W}}_{m}(i,j)^{2}x_{m}(j)^{2}\right)  + 2\sum E\left({\bf{W}}_{m}(i,j)x_{m}(j){\bf{W}}_{m}(k,l)x_{m}(l)\right). \\ \enum{in203}
\end{split}
\ee 
From proposition 1, ${\bf{W}}_{m}(i,j)$ and ${\bf{W}}_{m}(k,l)$ will (a) be independent of each other, and (b) be independent of $x_{m}(j)x_{m}(l)$. Using these results, along with the fact that $E({\bf{W}}_{m}(i,j)) = 0 $ gives 
\newline
\be 
E\left({\bf{W}}_{m}(i,j)x_{m}(j){\bf{W}}_{m}(k,l)x_{m}(l)\right) = E({\bf{W}}_{m}(i,j))E({\bf{W}}_{m}(k,l))E(x_{m}(j)x_{m}(l)) = 0. \enum{in2041}
\ee
\newline
Plugging \eref{in2041} into \eref{in203} gives
\be 
\begin{split}
u_{m}^{2} &= \sum\limits_{j=1}^{j=N} E\left({\bf{W}}_{m}(i,j)^{2}x_{m}(j)^{2}\right) \\ 
&= \sum\limits_{j=1}^{j=N} E\left({\bf{W}}_{m}(i,j)^{2}\right)E\left(x_{m}(j)^{2}\right) \\  
&= Nv^{2}(s_{m}^{2} + \mu_{m}^{2}) \\ \enum{in204}
\end{split}
\ee
for all $i$. Interestingly, this result holds for any arbitrary covariance structure of the inputs to the $m^{th}$ layer. 

Equations \eref{in201} and \eref{in204} provide insights into the central moments of $y_{m}(i)$, but can we derive insights into the distribution of $y_{m}(i)$? To answer this question, I make the additional assumption that the number of nodes in the hidden layer ($N$) is `large'; this assumption is reasonable given that most neural networks have several hundred nodes in the hidden layers. Under this assumption, we have the following result 

\begin{prop}
$y_{m}(i)$ will be approximately normally distributed for all values of $m$ and $i$.
\end{prop}
\noindent
\noindent
For the first iteration, note that $E(p_{ij}) = E\left({\bf{W}}_{m}(i,j)x_{m}(j)\right) = E\left({\bf{W}}_{m}(i,j)\right)E\left(x_{m}(j)\right)  = 0$. Furthermore, for $j \neq k$,
\be 
\begin{split}
Cov(p_{ij},p_{ik})&= E(p_{ij}p_{ik}) - E(p_{ij})E(p_{ik}) \\
&=  E(p_{ij}p_{ik}) \\
&= E\left({\bf{W}}_{m}(i,j)x_{m}(j){\bf{W}}_{m}(i,k)x_{m}(k)\right) = 0,\\ 
\end{split}
\ee
where the last equality follows from \eref{in2041}. This implies that $p_{i1}$, $p_{i2}$ ... $p_{iN}$ are independent and identically distributed random variables. Therefore by the Central Limit Theorem, we expect $y_{m}(i) = \sum\limits_{j=1}^{j=N} p_{ij}$ to converge to a normal distribution when $N$ is large \cite{lehmann2004elements}.
Even when $p_{i1}$, $p_{i2}$ ... $p_{iN}$ are dependent and not identically distributed, the conditions required to ensure that  $y_{m}(i) = \sum\limits_{j=1}^{j=N} p_{ij}$ converges to a normal distribution are weak (for a list of all the conditions, see Theorem 2.8.2 in \cite{lehmann2004elements}). Thus,  $y_{m}(i)$ is expected to  to be approximately normally distributed  during most iterations.
\qedsymbol 
\linebreak
\newline
\noindent
The analysis thus far has focused on providing general insights into the distribution of ${y}_{m}(i)$ resulting from equation \eref{in1}. In order to analyze the role of the non-linearity induced by \eref{in2}, assumptions need to be made about the nature of $g(x)$. In particular, my analysis critically hinges on the differentiability of $g(x)$ at 0. Accordingly, I split the analysis into two cases. The first case deals with the general class of activation functions differentiable at 0. In the second case, instead of considering all possible non-differentiable functions, I focus on the Rectified Linear Unit (RELU) which is commonly used in the analysis of neural networks.
\section{Activation functions differentiable at 0}
When $g(x)$ is differentiable at 0, we can perform a Taylor expansion in \eref{in2} about $E(y_{m}(i)) = 0$. Assuming that the higher order terms can be ignored, 
\be 
x_{m+1}(i) \approx g(0) + (y_{m} - 0)g'(0). \enum{in205}
\ee 
Taking the expectation in \eref{in205} gives
\be 
\mu_{m+1} = E(x_{m+1}(i)) \approx g(0). \enum{in206}
\ee
This equation suggests that the expected value of the inputs to the $(m+1)^{th}$ layer has little dependence on the moments of the inputs to the ${m}^{th}$ layer. Using this result recursively suggests that for all layers (barring the first), 
\be 
\mu_{j} = g(0) \mbox{ for all } j\geq 1 \enum{in2061}.
\ee 
Using \eref{in204} and \eref{in2061}, the variance of $x_{m+1}(i)$ can be computed from \eref{in205} as 
\be 
\begin{split}
s_{m+1}^{2} = Var(x_{m+1}(i)) &\approx (g'(0))^{2} Var(y_{m}(i)) \\ 
&= N(g'(0))^{2}v^{2}(s_{m}^{2} + \mu_{m}^{2}) \\ 
&= N(g'(0))^{2}v^{2}(s_{m}^{2} + g(0)^{2}) . \enum{in207}
\end{split}
\ee 
Using the condition $s_{m}^{2} \approx s_{m-1}^{2} ... = s_{1}^{2} = 1$ along with \eref{in2061} and \eref{in207} gives
\be 
v^{2} = \frac{1}{N(g'(0))^{2}(1+g(0)^{2})}, \enum{in208}
\ee 


\noindent
Equation \eref{in208} provide a general weight initialization strategy for any arbitrary differentiable activation function. I use the results developed in this section to analyze the optimal value of $v^{2}$ for two commonly used differentiable activation functions - the hyperbolic tangent and the sigmoid.\footnote{In their calculations, \cite{glorot2010understanding} and \cite{he2015delving} impose an additional set of constraints to ensure that the variance is maintained even during the backward pass. I believe that this is not required, since the requirement that the variance of the inputs at each layer be the same ensures that the gradient flows through in the backward pass.} 
\begin{figure}
\centering
\includegraphics[width=0.8\textwidth]{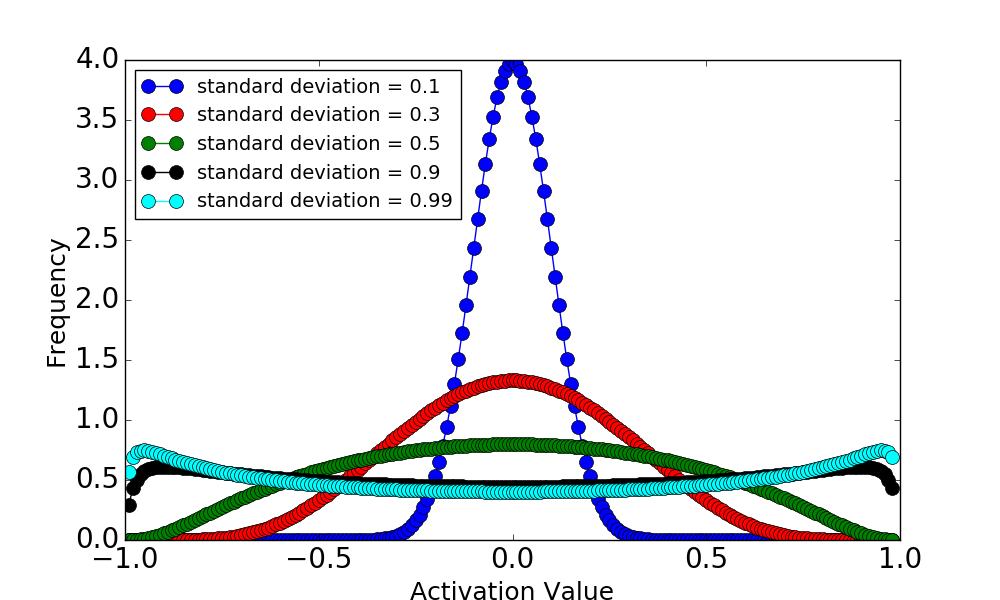}
\caption{\label{fig:tanh}Plots of the pdf described in \eref{tanh1} for different values of the standard deviation ($u_{m}$) of $y_{m}(i)$.}
\end{figure}

\subsection{Hyperbolic tangent activation}
For the hyperbolic tangent 
\be 
g(x) =\mbox{tanh}(x),
\ee
we have $g(0) = 0$ and $g'(0) = 1$. Plugging these results in \eref{in208} yields
\be 
v \approx \frac{1}{\sqrt{N}}, \enum{tanh1}
\ee 
which is precisely the {\it{Xavier Initialization}}.

\subsection*{Sequential saturation with the hyperbolic tangent}

In their analysis of a neural network with the hyperbolic tangent activations, \cite{glorot2010understanding}  find that the deeper layers in the neural network have a greater proportion of unsaturated nodes than the shallower layers. As is stated in their paper, `why this is happening remains to be understood'.

To explain their finding, I begin by noting that in  \cite{glorot2010understanding}, the authors initialize the weights using samples from a uniform distribution $\left(U[-1/\sqrt{N},1/\sqrt{N}]\right)$ having a variance of  $1/3N$. Therefore, from \eref{in2061} and \eref{in207}, with $g(0) = 0$ and $g'(0) = 1$, we have $\mu_{m} = 0$ and



\be 
s_{m+1}^{2} = \frac{1}{3}s_{m}^{2} < s_{m}^{2}
\ee \enum{seq1}
respectively. From \eref{in204}, this implies that $u_{m+1}^{2}<u_{m}^{2}$ for all $m$ (i.e., $u_{m}^{2}$ is a decreasing function of $m$).  Furthermore from proposition 2, $y_{m}(i) \sim {\cal{N}}(0,u_{m}^{2})$. Therefore, $x_{m+1}(i)$ will be the tanh transformation of a normal random variable. Using results from \cite{godfrey2009tanh}, $x_{m+1}(i) = \mbox{tanh}(y_{m}(i))$ will have a probability density function (pdf) given by 
 
\be 
f(y) = \frac{1}{1-y^{2}}\frac{1}{\sqrt{2\pi u_{m}^{2}}}\mbox{e}^{\left({-\frac{t_{y}^{2}}{2u_{m}^{2}}}\right)} \enum{tanh1},
\ee
where 
\be 
t_{y} =  \frac{1}{2} ln\left(\frac{1+y}{1-y}\right).
\ee
Plots of this pdf for different values of $u_{m}$  (provided in Figure \ref{fig:tanh}) produce trends similar to those observed by the simulation studies of \cite{glorot2010understanding} (figure 4 in their paper). A comparison of Figure \ref{fig:tanh} and figure 4 of \cite{glorot2010understanding}  suggests that $u_{m}^{2}$ is a decreasing function of $m$, as is expected.

From the results in \cite{godfrey2009tanh}, we expect the  activations to be (a) approximately normally distributed when $u_{m}$ is close to 0 and (b) bimodally distributed with local maximas near -1 and +1 when $u_{m}$ is close to 1. Accordingly, since $u_{m}^{2}$ is a decreasing function of $m$, we expect the activations from the shallower layers to be more saturated (i.e., more concentrated near -1 and +1), and the saturation in the activations to reduce as we go to the deeper layers in the network. 

\subsection{Sigmoid activation}

\begin{figure}
\centering
\begin{minipage}{.3\textwidth}
  \centering
  \includegraphics[width=.5\linewidth]{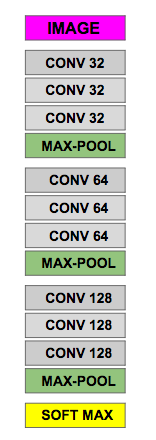}
  \captionof{figure}{Architecture of deep nural network used in analysis of sigmoid; stride lengths in all layers are $2 \times 2$.}
  \label{fig:nnarch}
\end{minipage}%
\hfill
\begin{minipage}{.6\textwidth}
  \centering
  \includegraphics[width=1\linewidth]{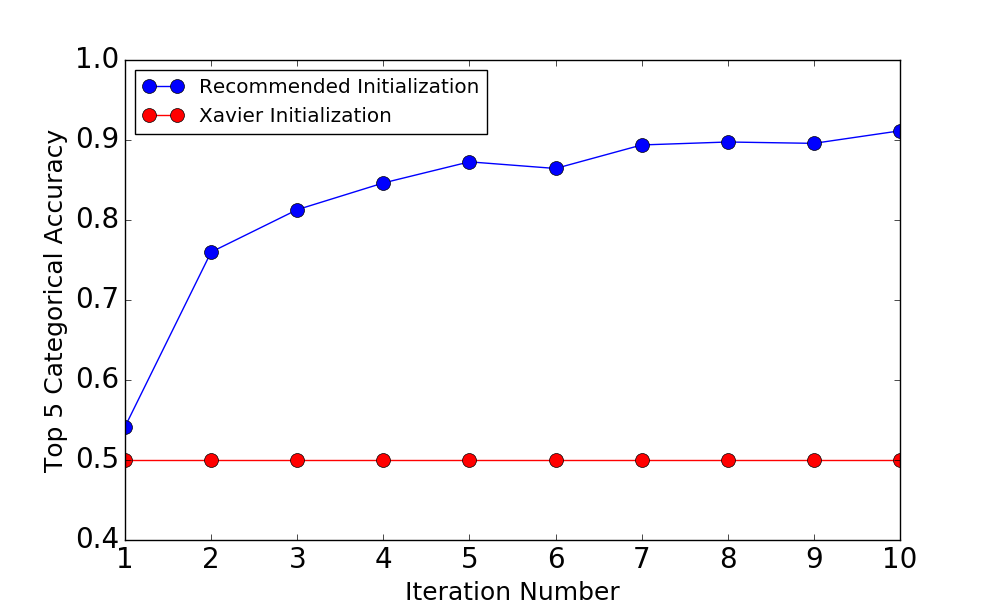}
  \captionof{figure}{Convergence comparison of {\it{Xavier initialization}} with initialization recommended in \eref{relur}. The {\it{Xavier initialization}} stalls while the initialization recommended in \eref{relur} converges proceeds rapidly towards convergence.}
  \label{fig:conv}
\end{minipage}
\end{figure}

For the sigmoid activation defined as
\be 
g(x) = \frac{1}{1 + \mbox{e}^{-x}}
\ee
we have $g(0) = 0.5$, $g'(0) = 1/4$. Plugging these values in \eref{in208} yields
\be 
v \approx \frac{3.6}{\sqrt{N}}, \enum{relur}
\ee 
To compare the initialization described in \eref{relur} with
the {\it{Xavier initialization}}, I use a simple 10 layer neural network whose architecture is described in Figure \ref{fig:nnarch}. For my experiments, I use the  CIFAR 10 dataset \cite{krizhevsky2009learning} comprising 60,000,  $32 \times 32$ color images evenly split over 10 classes. The dataset comprises 50,000 training examples (which forms the training dataset in my analyses) and 10,000 test examples (which forms the validation dataset in my analyses). 

First, I train the neural network with the {\it{Xavier initialization}} for 10 iterations and compute the top 5 accuracy on the validation dataset for each iteration. Next, I repeat the process using the initialization stated in \eref{relur}. A comparison of the validation accuracies for the 2 cases is provided in Figure \ref{fig:conv}, which shows that the convergence appears to stall with the {\it{Xavier initialization}}, but proceeds rapidly with the initialization proposed in \eref{relur}.\footnote{Python code (using the package Keras \cite{chollet2015}) to replicate Figure \ref{fig:conv} can be downloaded from \url{https://github.com/sidkk86/weight_initialization}}

\section{Activation functions not differentiable at 0}
When $g(x)$ is not differentiable at 0, the analysis seems more difficult than in the previous section. Instead of attempting to provide a general solution, I focus on the most important non-differentiable activation function used in the analysis of neural networks - the Rectified Linear Unit (RELU).

\subsection{RELU activation}

Since the RELU activation is not differentiable at 0, the results from \erefs{in205}{in207} cannot be used to compute the optimal value of $v^{2}$. To proceed, I use proposition 2 and \eref{in201} which state that for the first iteration, $ y_{m}(i) \sim {\cal{N}}(0,u_{m}^{2})$. We are interested in in the mean and variance of $x_{m+1}(i) = \mbox{max}(0,y_{m}(i))$. The mean will be given by
\be 
 \mu_{m+1} = E(y_{m}(i){\bf{I}}(y_{m}(i)>0)) = \frac{1}{{u_{m}\sqrt{2\pi}}}\int_{0}^{\infty} x \mbox{e}^{-\frac{x^{2}}{2u_{m}^{2}}}dx = \frac{u_{m}}{\sqrt{2\pi}}\int_{0}^{\infty}\mbox{e}^{-t}dt = \frac{u_{m}}{\sqrt{2\pi}}. \enum{relu1}
\ee 
Similarly,
\be 
E(x_{m+1}(i)^{2}) =  \frac{1}{{u_{m}\sqrt{2\pi}}}\int_{0}^{\infty} x^{2} \mbox{e}^{-\frac{x^{2}}{2u_{m}^{2}}}dx = \left(\frac{1}{2}\right) \frac{1}{{u\sqrt{2\pi}}} \int_{-\infty}^{\infty} x^{2} \mbox{e}^{-\frac{x^{2}}{2u^{2}}}dx = \frac{u_{m}^{2}}{2}. \enum{relu2}
\ee 
Using \eref{relu1} and \eref{relu2},
\be
s_{m+1}^{2} = E(x_{m+1}(i)^{2}) - \mu_{m+1}^{2} = \frac{u_{m}^{2}}{2} - \frac{u_{m}^{2}}{2\pi} \approx 0.34u_{m}^{2}. \enum{relu3}
\ee 
For the variance to be maintained at each iteration, we require $s_{m+1}^{2}  \approx 1$ which yields 
\be 
u_{m}^{2} \approx 3. \enum{relu301}
\ee 
Plugging \eref{relu301} in \eref{relu1} yields $\mu_{m+1} \approx 0.7$. By the symmetry of the problem during the first iteration, we expect $\mu_{m+1} \approx \mu_{m} ... \approx \mu_{2} \approx 0.7$. 
Using this result in \eref{in204} yields
\be 
3 = Nv^{2}(1+ 0.49) 
\ee 
or 
\be 
v^{2}\approx 2/N,
\ee 
which is consistent with that obtained by \cite{he2015delving}.
\subsubsection*{To converge or not to converge, that is the question.}

In \cite{he2015delving} paper, the authors provide an example of a 22 layer neural network using RELU activations which converges with the {\it{Xavier Initialization}}, and a 30 layer neural network which does not converge with the same initializations and activation functions. 

\begin{figure}
\centering
\includegraphics[width=1\textwidth]{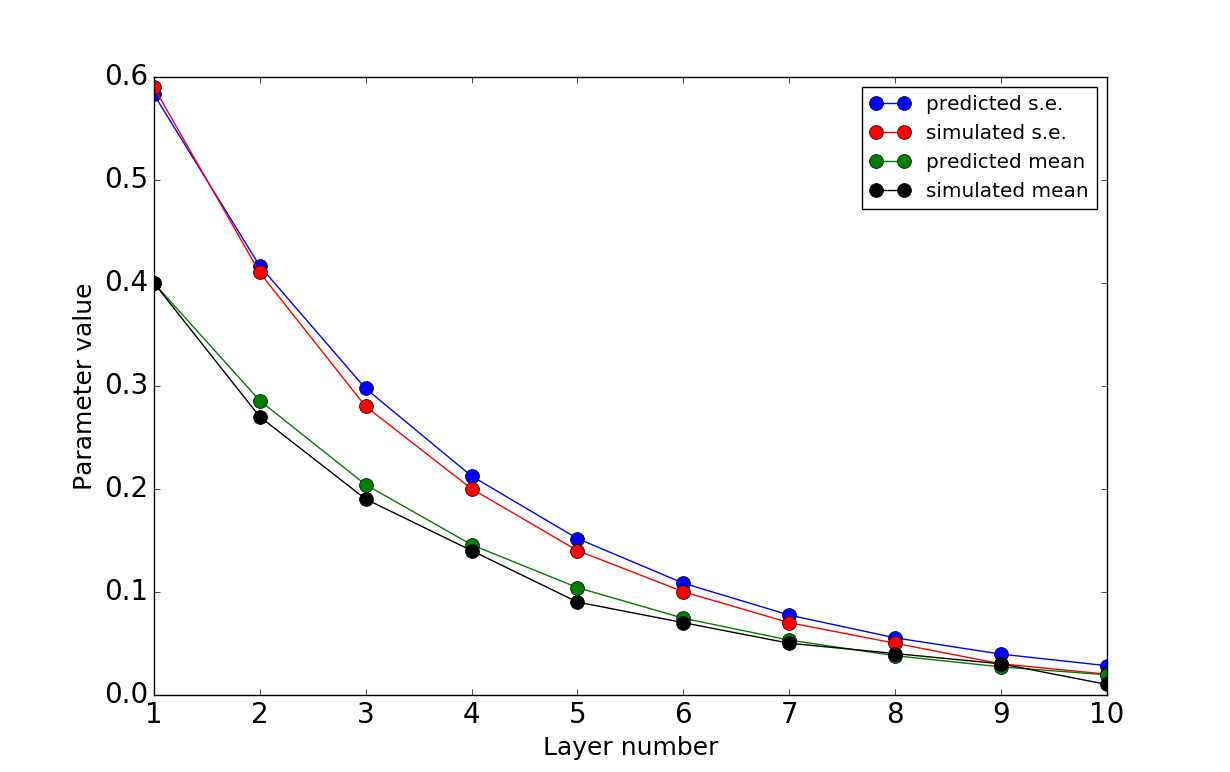}
\caption{\label{fig:cs231ncomp}A comparison of the predicted means and standard errors obtained from  \erefs{relu603}{relu602} with the simulated values reported  in slide 61 of \cite{cs231nlecture5}.}
\end{figure}

To understand why this happens, I compute the central moments of $x_{m+1}$ in terms of the central moments of $x_{m}$ when $v^{2} = 1/N$. From \eref{in204} we have
\be 
u_{m}^{2} = s_{m}^{2} + \mu_{m}^{2}. \enum{r01}
\ee
Plugging results from \eref{r01} into \eref{relu1} yields the recursion
\be 
\mu_{m+1}^{2} = \frac{1}{2\pi}(s_{m}^{2} + \mu_{m}^{2}) \approx 0.16(s_{m}^{2} + \mu_{m}^{2}). \enum{r02}
\ee
Similarly, plugging results from \eref{r01} and \eref{r02} in \eref{relu3} yields
\be 
s_{m+1}^{2}  \approx 0.34(s_{m}^{2} + \mu_{m}^{2}). \enum{relu5}
\ee
Simple manipulations of equations \erefs{r01}{relu5} gives
\be 
\mu_{m}^{2} \approx 0.16 \times (0.51)^{m-1} \enum{relu603}
\ee
and
\be 
s_{m}^{2} \approx 0.34 \times (0.51)^{m-1}. \enum{relu602}
\ee
for all $m \geq 1$. These approximations are remarkably accurate, as is demonstrated by comparisons with the simulation experiments described in Figure \ref{fig:cs231ncomp}. 

Equation \eref{relu602} shows that the variance of the inputs to the deeper layers is exponentially smaller than the variance of the inputs to the shallower layer. Therefore, the deeper the neural network, the worse the performance of the {\it{Xavier Initialization}} will be. From \eref{relu602}, $s_{22}^{2} \approx 1.62 \times 10^{-7}$ and $s_{30}^{2} \approx 6.33 \times  10^{-10}$. Thus the variance in the input to the $30^{th}$ layer will be $(0.51)^{8} = 3 \times 10^{-3}$ times smaller than the variance to the $22^{nd}$ layer, and explains the possible reason why the 30 layer neural network described in \cite{he2015delving} converges, but the 22 layer neural network does not. \footnote{It is surprising that the 22 layer neural network converges!}

\section{Conclusion}
In this paper, I have provided a general framework for weight initialization with non-linear activation functions. First, I provide a general formula for the ideal weight initialization for all activation functions differentiable at 0. I show how the weight initializations change for the hyperbolic tangent and sigmoid activation functions. Second, I focus only on the Rectified Linear Unit (RELU) from the class of functions that are non-differentiable at 0, and I provide a rigorous proof of the {\it{He Initialization}}. Finally, I show why the {\it{Xavier initialization}} fails to work with the RELU activation function. Given the sharp increase in non-differentiable activation functions over the years, a more general version of my (largely incomplete) analysis of non-differentiable functions is warranted. My analysis repeatedly illustrates the drastic difference in dynamics which can result from introducing non-linearities in the system.



\bibliographystyle{plain}
\bibliographystyle{plain}

\begin{thebibliography}{10}

\bibitem{chollet2015}
François Chollet.
\newblock Keras.
\newblock \url{https://github.com/fchollet/keras}, 2015.

\bibitem{collobert2008unified}
Ronan Collobert and Jason Weston.
\newblock A unified architecture for natural language processing: Deep neural
  networks with multitask learning.
\newblock In {\em Proceedings of the 25th international conference on Machine
  learning}, pages 160--167. ACM, 2008.

\bibitem{glorot2010understanding}
Xavier Glorot and Yoshua Bengio.
\newblock Understanding the difficulty of training deep feedforward neural
  networks.
\newblock In {\em Aistats}, volume~9, pages 249--256, 2010.

\bibitem{godfrey2009tanh}
Michael~D Godfrey.
\newblock The tanh transformation.
\newblock {\em Information Systems Laboratory, Stanford University}, 2009.

\bibitem{graves2013speech}
Alex Graves, Abdel-rahman Mohamed, and Geoffrey Hinton.
\newblock Speech recognition with deep recurrent neural networks.
\newblock In {\em Acoustics, speech and signal processing (icassp), 2013 ieee
  international conference on}, pages 6645--6649. IEEE, 2013.

\bibitem{he2015delving}
Kaiming He, Xiangyu Zhang, Shaoqing Ren, and Jian Sun.
\newblock Delving deep into rectifiers: Surpassing human-level performance on
  imagenet classification.
\newblock In {\em Proceedings of the IEEE international conference on computer
  vision}, pages 1026--1034, 2015.

\bibitem{cs231nlecture5}
Andrej Karpathy, Justin Johnson, and Fei~Fei Li.
\newblock Cs 231n: Convolutional neural networks for visual recognition,
  lecture 5, slide 61.
\newblock \url{http://cs231n.stanford.edu/slides/2016/winter1516_lecture5.pdf},
  2016.

\bibitem{kim2014convolutional}
Yoon Kim.
\newblock Convolutional neural networks for sentence classification.
\newblock {\em arXiv preprint arXiv:1408.5882}, 2014.

\bibitem{krizhevsky2009learning}
Alex Krizhevsky and Geoffrey Hinton.
\newblock Learning multiple layers of features from tiny images.
\newblock 2009.

\bibitem{lehmann2004elements}
Erich~Leo Lehmann.
\newblock {\em Elements of large-sample theory}.
\newblock Springer Science \& Business Media, 2004.

\bibitem{maas2013rectifier}
Andrew~L Maas, Awni~Y Hannun, and Andrew~Y Ng.
\newblock Rectifier nonlinearities improve neural network acoustic models.
\newblock In {\em Proc. ICML}, volume~30, 2013.

\bibitem{mishkin2015all}
Dmytro Mishkin and Jiri Matas.
\newblock All you need is a good init.
\newblock {\em arXiv preprint arXiv:1511.06422}, 2015.

\bibitem{russakovsky2015imagenet}
Olga Russakovsky, Jia Deng, Hao Su, Jonathan Krause, Sanjeev Satheesh, Sean Ma,
  Zhiheng Huang, Andrej Karpathy, Aditya Khosla, Michael Bernstein, et~al.
\newblock Imagenet large scale visual recognition challenge.
\newblock {\em International Journal of Computer Vision}, 115(3):211--252,
  2015.

\bibitem{saxe2013exact}
Andrew~M Saxe, James~L McClelland, and Surya Ganguli.
\newblock Exact solutions to the nonlinear dynamics of learning in deep linear
  neural networks.
\newblock {\em arXiv preprint arXiv:1312.6120}, 2013.

\bibitem{socher2011parsing}
Richard Socher, Cliff~C Lin, Chris Manning, and Andrew~Y Ng.
\newblock Parsing natural scenes and natural language with recursive neural
  networks.
\newblock In {\em Proceedings of the 28th international conference on machine
  learning (ICML-11)}, pages 129--136, 2011.

\bibitem{sussillo2014random}
David Sussillo and LF~Abbott.
\newblock Random walk initialization for training very deep feedforward
  networks.
\newblock {\em arXiv preprint arXiv:1412.6558}, 2014.

\bibitem{sutskever2014sequence}
Ilya Sutskever, Oriol Vinyals, and Quoc~V Le.
\newblock Sequence to sequence learning with neural networks.
\newblock In {\em Advances in neural information processing systems}, pages
  3104--3112, 2014.

\bibitem{szegedy2013deep}
Christian Szegedy, Alexander Toshev, and Dumitru Erhan.
\newblock Deep neural networks for object detection.
\newblock In {\em Advances in Neural Information Processing Systems}, pages
  2553--2561, 2013.

\bibitem{toshev2014deeppose}
Alexander Toshev and Christian Szegedy.
\newblock Deeppose: Human pose estimation via deep neural networks.
\newblock In {\em Proceedings of the IEEE Conference on Computer Vision and
  Pattern Recognition}, pages 1653--1660, 2014.

\end{thebibliography}

\end{document}